%% file: sample_FG2024.tex
\documentclass[a4paper, 10pt, conference]{ieeeconf}      
\let\labelindent\relax

\usepackage{FG2024}
\usepackage{times}
\usepackage{epsfig}
\usepackage{graphicx}
\usepackage{amsmath}
\usepackage{amssymb}
\usepackage{tabularx}
\usepackage[pagebackref]{hyperref}
\usepackage{import}
\usepackage{stfloats}
\usepackage{verbatim}
\usepackage{xcolor,colortbl}
\usepackage[section]{placeins}
\usepackage{subcaption}
\usepackage{balance}
\usepackage{bm}
\usepackage{soul}
\usepackage{enumitem}
\usepackage{wrapfig}

\definecolor{blue}{RGB}{0,0,0}
\definecolor{boxgrey}{HTML}{F3F3F3}
\newcolumntype{a}{>{\columncolor{boxgrey}}r}

\FGfinalcopy 

\IEEEoverridecommandlockouts                              
\def\FGPaperID{63} 

\title{\LARGE \bf
If It’s Not Enough, Make It So: Reducing Authentic Data Demand in Face Recognition through Synthetic Faces}

\usepackage{fancyhdr}
\begin{document}

\ifFGfinal
\author{\parbox{16cm}{\centering
    {\large Andrea Atzori$^1$, Fadi Boutros$^{2,3}$, Naser Damer$^{2,3}$, Gianni Fenu$^1$, Mirko Marras$^1$}\\
    {\normalsize
    $^1$ Department of Mathematics and Computer Science, University of Cagliari, Cagliari, Italy\\
    $^2$ Fraunhofer Institute for Computer Graphics Research, Germany\\
    $^3$ Department of Computer Science, TU Darmstadt, Darmstadt, Germany}}%
    \thanks{We acknowledge financial support under the National Recovery and Resilience Plan (NRRP), Mission 4 Component 2 Investment 1.5 - Call for tender No.3277 published on December 30, 2021 by the Italian Ministry of University and Research (MUR) funded by the European Union – NextGenerationEU. Project Code ECS0000038 – Project Title eINS Ecosystem of Innovation for Next Generation Sardinia – CUP F53C22000430001- Grant Assignment Decree No. 1056 adopted on June 23, 2022 by the Italian Ministry of University and Research. This work has been also funded by the German Federal Ministry of Education and Research and the Hessen State Ministry for Higher Education, Research and the Arts within their joint support of the National Research Center for Applied Cybersecurity ATHENE.}
}
\else
\author{Anonymous FG2024 submission\\ Paper ID \FGPaperID \\}
\pagestyle{plain}
\fi
\maketitle

\thispagestyle{fancy}

\begin{abstract}
Recent advances in deep face recognition have spurred a growing demand for large, diverse, and manually annotated face datasets. Acquiring authentic, high-quality data for face recognition has proven to be a challenge, primarily due to privacy concerns. Large face datasets are primarily sourced from web-based images, lacking explicit user consent. In this paper, we examine whether and how synthetic face data can be used to train effective face recognition models with reduced reliance on authentic images, and thus mitigating large authentic data collection concerns. First, we explored the performance gap among recent state-of-the-art face recognition models, trained only on synthetic data or authentic data. Then, we deepened our analysis by training a state-of-the-art backbone with various combinations of synthetic and authentic data, gaining insights into optimizing the limited use of the latter for verification accuracy. Finally, we assessed the effectiveness of data augmentation approaches on synthetic and authentic data, with the same goal in mind. Our results highlighted the effectiveness of FR trained on combined datasets, particularly when combined with appropriate augmentation techniques.
\end{abstract}

\vspace{1mm}
\section{Introduction}
Face Recognition (FR) stands as one of the predominant biometric technologies in various applications, ranging from logical access control~\cite{adaface, cosface, elasticface} to portable devices~\cite{pocketnet}. Existing FR models have reached impressive accuracy thanks to the recent advances in deep learning, including margin-penalty-based softmax losses~\cite{arcface, cosface, elasticface, adaface} and neural architectures~\cite{residualFR, unsupervisedFR}. This development, however, has been made possible by the public availability of large-sized, identity-labeled, FR training databases~\cite{VGGFace2, msceleb1m, webface260m}.

\begin{figure}[ht!]
\centering
\includegraphics[width=1.\linewidth]{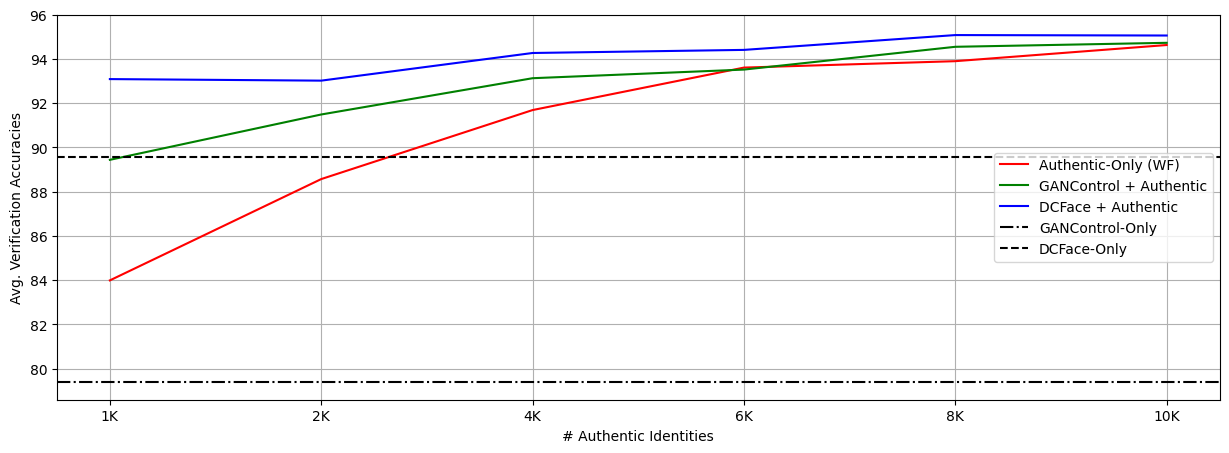}
\caption{{\color{blue} Average verification accuracy on the testing dataset (y-axis) vs. number of authentic identities from CASIA-WebFace (WF) \cite{latentofromscratch} in the training dataset (x-axis). In all settings, the verification accuracy improved by increasing the number of authentic identities in the training dataset. Also, combining synthetic (black line: DCFace \cite{DCFace}, green line: ExFaceGAN(GANControl) \cite{ExFaceGAN}) with a limited subset of authentic data improved the FR performance, in comparison to the case where only a limited subset of authentic identities (red line) is used to train FR.  
The number of synthetic identities in the combined dataset experiments is fixed (10K). 
}}
\vspace{-3mm}
\label{fig:RQ1casiaavg}
\end{figure}

Face databases include millions of images often gathered from the Internet without user consent, thereby giving rise to concerns about their legal and ethical use for FR development. Notably, the approval of the General Data Protection Regulation (GDPR)~\cite{GDPR} by the European Union (EU) in 2018 has significantly amplified criticism concerning the privacy-related issues associated with the use of publicly available face datasets obtained without adequate consent. The GDPR explicitly affirms individuals' "right to be forgotten" and imposes more stringent regulations concerning the acquisition, distribution, and utilization of biometric data, making adherence to such regulations a key, intricate endeavor when applied to face datasets. Consequently, several databases frequently employed for FR training, including but not limited to~\cite{VGGFace2, umdfaces, msceleb1m}, have been retracted by their creators, to prevent potential legal issues. These circumstances have raised doubts about the continuity of FR research, as the availability of such fundamental resources has become uncertain.

Synthetic data has recently emerged as a promising alternative to genuine datasets for FR training~\cite{DBLP:journals/ivc/BoutrosSFD23}, serving as a viable solution to address the prevailing legal and ethical concerns~\cite{SynFace, Sface, unsupervisedFR, Digiface1m,DBLP:conf/cvpr/KolfREBKD23}. This research trajectory has been significantly propelled by the advancement of Deep Generative Models (DGMs), which facilitate the generation of synthetic samples thanks to their ability to capture the probability distribution underlying a face dataset. This generative process can be constrained to various factors, including but not limited to age, facial expression, head pose, and lighting~\cite{disentangled, GANControl, Stylerig, GIF}. The majority of DGM techniques employed for the creation of synthetic faces build upon Generative Adversarial Networks (GANs)~\cite{GANs, disentangled, latentinterp, GANControl, Sface} or Diffusion Models (DMs)~\cite{denoising, improveddiff, diffgan}. 

Deep face recognition methods, such as~\cite{GANassess, unsupervisedFR, SynFace}, have increasingly leveraged GANs~\cite{disentangled, limitedGANs} to generate synthetic data for FR training. However, state-of-the-art (SOTA) FR models trained on synthetic data have often exhibited a drop in verification accuracy, in comparison to FR models trained on authentic data. This discrepancy in performance primarily stems from the training datasets' limited identity discrimination~\cite{Sface} or their minimal intra-class variance~\cite{unsupervisedFR, SynFace}. DMs have emerged as a viable alternative to GANs for image synthesis, but this progress comes at the cost of stability and a substantial reduction in training performance. Unfortunately, several questions are still open on the extent to which authentic and synthetic data can be effectively combined to tackle the limitations inherent to both domains. 

In this paper, our primary objective is to investigate the suitability of combined (authentic and synthetic) training datasets for model development in FR. This exploration can provide a better understanding of how to address both the performance inadequacies associated with synthetic datasets and the privacy-related concerns that characterize authentic datasets, thereby facilitating the creation of highly accurate FR models, with reduced authentic data demand. As teased in Figure \ref{fig:RQ1casiaavg}, training FR models with a combination of synthetic data and subsets of authentic identities is visibly beneficial in terms of FR performance in comparison to cases where only limited subsets of authentic identities are available, as we assumed that the synthetic data could be generated without any limitation and that small numbers of authentic identities could be collected with proper user consent. Specifically, our novel contribution is threefold:
\begin{itemize}
    \item {\color{blue} We investigated the impact of training models with different ratios of synthetic and authentic identities, under a fixed total number of training images.}
    \item We explored whether a training dataset with a fixed amount of synthetic identities and an increasing number of authentic ones could lead to (at least) comparable accuracy w.r.t. a large, authentic-only dataset.
    \item We evaluated the influence of training data augmentation on accuracy, by augmenting (i) both the synthetic and authentic subsets and (ii) exclusively the synthetic subset, identifying and discussing the optimal settings.
\end{itemize}

\section{Related Work}
With the ongoing success of deep generative models in generating high-quality and realistic face images~\cite{GANControl,DiscoFaceGAN,denoising,improveddiff}, several recent works proposed the use of synthetic data in FR development~\cite{SynFace,unsupervisedFR,Digiface1m,Sface,IDiff-Face,DCFace,DBLP:journals/ivc/BoutrosSFD23}.
The synthetic data for FR training can be categorized based on the type of underlying generative model under three categories, GAN-Based~\cite{SynFace,unsupervisedFR,Sface,ExFaceGAN}, digital rendering~\cite{Digiface1m}, and diffusion-based~\cite{GANDiffFace,IDiff-Face,DCFace}

SynFace~\cite{SynFace} and SFace~\cite{Sface} were among the earliest works that proposed the use of GAN-generated face images for FR training. 
To enhance intra-class diversity in synthetic data, SynFace proposed to generate synthetic data using an attribute-conditional GAN model, i.e., DiscoFaceGAN~\cite{DiscoFaceGAN}, and perform identity and domain mixup.
SFace also analyzed the impact of Style-GAN~\cite{DBLP:conf/cvpr/KarrasLAHLA20} training under class conditional settings. Furthermore, the extent to which transferring knowledge from the pretrained model on authentic data improves the performance of synthetic-based FR was analyzed, going beyond FR training only on synthetic data. 
UsynthFace~\cite{unsupervisedFR} leveraged unlabelled synthetic data for unsupervised FR training. The work also presented an extensive data augmentation method based on GANs~\cite{DiscoFaceGAN} along with color and geometric transformation~\cite{randaug}.
Conversely, ExFaceGAN~\cite{ExFaceGAN} presented a framework to disentangle identity information in learned latent spaces of unconditional GANs, aiming at generating multiple samples of any synthetic identity. 

DigiFace-1M~\cite{Digiface1m} employed a digital rendering pipeline, leveraging facial geometry models, a diverse array of textures, hairstyles, and 3D accessories, along with robust data augmentation techniques during training.  However, it comes at a considerable computational cost during the rendering process. DigiFace-1M also proposed to combine synthetic with authentic data during FR training to improve the verification accuracy of synthetic-based FR using a small and fixed number of authentic identities.

Very recently, IDiff-Face~\cite{IDiff-Face} and DCFace~\cite{DCFace} adopted diffusion models to generate synthetic data for FR training, achieving SOTA verification accuracy for synthetic-based FR. Specifically, IDiff-Face included fizziness in the identity condition to induce variations in the generated data. DCFace proposed a two-stage generative framework. In the first stage, an image of a novel identity using an unconditional diffusion model is generated and an image style from the style bank is selected. In the second stage, the generated images and style are mixed using a dual conditional diffusion model.
GANDiffFace~\cite{GANDiffFace} proposed to combine GAN and diffusion models to generate synthetic data for FR training that exhibit certain attributes such as controllable demographic distribution. 

{\color{blue}Although these approaches~\cite{DCFace,IDiff-Face,Digiface1m,Sface,SynFace} achieved relatively high verification accuracy when they were evaluated on authentic datasets, their performance is lower than the one obtained by FR models trained on authentic data~\cite{arcface}.
To overcome this challenge, SFace~\cite{Sface} applied knowledge distillation on the embedding level between the model trained on the authentic data and the one trained on synthetic data.
DigiFace~\cite{Digiface1m} and DCFace~\cite{DCFace} combined their synthetic datasets with a subset of the authentic CASIA-WebFace dataset~\cite{latentofromscratch}, aiming at improving the verification accuracy of synthetic-based FR. Both methods demonstrated that synthetic-based verification accuracy can be improved using a limited set of authentic data.  Wang et at.~\cite{wang2023boosting} proposed to combine synthetic data from SynFace~\cite{SynFace} with a small and limited subset of authentic data. The work proposed to utilize the same backbone to extract features from synthetic and authentic identity and then learn multi-class classification through two classification layers for synthetic and authentic classes.
However, these works~\cite{Digiface1m,DCFace,wang2023boosting} only utilized a subset of authentic data, with a fixed number of authentic identities, and they did not investigate the performance gain when different ratios of synthetic and authentic identities combined in the FR training.
Recently, several challenges and competitions have been organized in conjunction with top venues, aiming at promoting privacy-friendly synthetic-based FR development. 
FRCSyn competition \cite{Melzi_2024_WACV} was organized at WACV 2024, aiming to explore the use of synthetic data in FR training and to attract the development of solutions for synthetic-based FR. The challenge considered two main tasks, training FR only with synthetic data and training FR with both synthetic and authentic data. The achieved results of the top-performing solutions from FRCSyn \cite{Melzi_2024_WACV} competition are further investigated and reported in \cite{MELZI2024102322}. A second edition of FRCSyn \cite{deandres2024frcsyn} was organized in conjunction with CVPR 2024.
Also, very recently SDFR  \cite{shahreza2024sdfr} competition was organized in conjunction with FG 2024, attracting developing and promoting solutions for synthetic-based FR.

Unlike previous research that combined fixed, (in terms of numbers of images and identities) sets of synthetic and authentic data~\cite{Digiface1m,DCFace,wang2023boosting,Melzi_2024_WACV}, we investigate in detail FR performances using different combinations of authentic and synthetic data.
We first fixed the total number of identities in the training dataset and investigated the performance of different ratios of synthetic and authentic identities. Then, we investigated the FR performance using a fixed number of synthetic identities with different subsets of authentic data. Finally, we investigated the impact of data augmentation on synthetic and authentic subsets.}

\section{Methodology}
This work aims, at first, to investigate the performance gaps between FR models trained on synthetic datasets and the ones trained on privacy-sensitive authentic datasets. Then, we explored the performance gains of synthetic-based FR by including
a limited set of authentic identities in the training dataset. Finally, we investigate the impact of data augmentation on synthetic-based FR verification accuracy. 



\subsection{Datasets Preparation}
The presented investigations in this paper are performed using four training datasets, two authentic and two synthetic. 

As the authentic data, we adopted the well-known CASIA-WebFace~\cite{latentofromscratch} {\color{blue}and MS1MV2~\cite{arcface}} datasets. 
CASIA-WebFace \cite{latentofromscratch} consists of 0.5M images of 10K identities. 
{\color{blue} MS1MV2 is the refined version of the MS-Celeb-1M dataset~\cite{msceleb1m}. MS1MV2 contains 5.8M images of 85K identities. 
To provide comparable results across FR trained on authentic-only data and on synthetic and authentic combinations, we utilized a subset of 10K identities of MS1MV2 in the conducted experiments. This subset will be noted as M2-S and contains the first 10K identities from MS1MV2.}

The synthetic datasets come from ExFaceGAN~\cite{ExFaceGAN} and DCFace~\cite{DCFace}. Each of these datasets contains 0.5M images of 10K identities, i.e., 50 images per identity. 
{\color{blue} The ExFaceGAN dataset is generated using an identity disentanglement approach \cite{ExFaceGAN} on pretrained GAN-Control~\cite{GANControl} which is trained on the FFHQ dataset~\cite{FFHQ}.}
{\color{blue} From now on, the ExFaceGAN(GAN-Control) dataset will be noted as GC.}

The second synthetic dataset is generated via DCFace~\cite{DCFace}. {\color{blue}The latter is based on two-stage diffusion models.
In the first stage, a high-quality face image of a novel identity is generated using unconditional diffusion models~\cite{denoising} trained on FFHQ~\cite{FFHQ}, with an image style randomly selected from a style bank. In the second stage, the generated images and styles from stage 1 are mixed using a dual conditional diffusion model~\cite{denoising} trained on CASIA-WebFace~\cite{latentofromscratch} to generate an image with a specific identity and style. 
The dataset generated by DCFace will be noted as DC.
All images are aligned and cropped to $112 \times 112$ using landmarks obtained by the Multi-task Cascaded Convolutional Networks (MTCNN)~\cite{zhang2016joint}, following~\cite{arcface}. All the training and
testing images are normalized to have values between
-1 and 1.}

\begin{figure*}[!t]
\centering
\includegraphics[width=.99\textwidth]{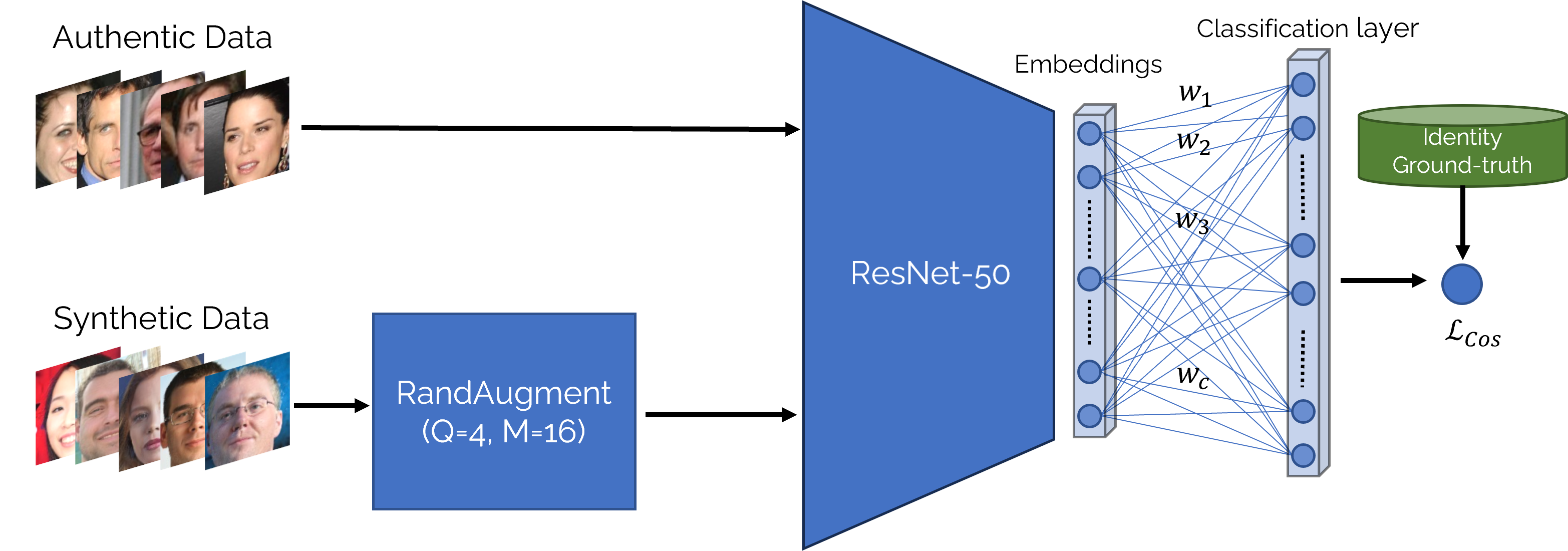}

\caption{{\color{blue}\textbf{FR Training Paradigm Overview}. Subsets of authentic and synthetic data are combined to form the training dataset. 
During the training phase, only synthetic data is augmented with RandAugment, as well discussed in Section \ref{sec:results}. the utilized network architecture in all settings is ResNet50~\cite{residualFR} trained with CosFace~\cite{cosface} loss.
}
}
\label{fig:pipeline}
\end{figure*}

{\color{blue}
\subsection{Data Mixing}
Given a synthetic dataset $S$ with a number of identities equal to $N$ and an authentic dataset $A$ with a number of identities equal to $M$, we combined these two datasets into one dataset $C = s \bigcup a$, where $s \subseteq S$ and $a \subseteq A$ are subsets of $S$ and $A$, respectively, with a total number of identities equal to $H$, as follows. In the first evaluation setting, we set $H$ to be equal to $M$ and sampled $m < M$ from the authentic dataset and  $M-m < N$ from the synthetic dataset. We evaluated  several $m \in [1 \mathrm{K}, 2 \mathrm{K}, 3 \mathrm{K}, 4 \mathrm{K}, 5 \mathrm{K}, 6 \mathrm{K}, 7 \mathrm{K}, 8 \mathrm{K}, 9 \mathrm{K}]$. 
In this stage, we aim to boost the verification accuracy of synthetic-based FR by combining synthetic with authentic face images into one dataset. We fixed the total number of identities, in this case, to provide a fair comparison across the considered settings.
In the second stage, we considered the whole synthetic dataset $S$ and combined it with a subset $a \subseteq A$. The sampled $a$ contains a number of identities equal to $1 \mathrm{K}$, $2 \mathrm{K}$, $4 \mathrm{K}$, $6 \mathrm{K}$, $8 \mathrm{K}$ or $10 \mathrm{K}$. 
In this scenario, we assumed that the synthetic data could be generated without any limitation and that a small number of authentic identities could be collected with proper user consent. }


\subsection{Data Augmentation} \label{sec:daug}
Considering the constrained range of visual diversity in synthetic face data, in comparison to large and realistic appearance variations in authentic data~\cite{DBLP:journals/ivc/BoutrosSFD23,unsupervisedFR,DCFace}, we evaluated in this work the impact of augmenting the FR training dataset on FR verification accuracy, as follows.   
\noindent \paragraph{Baseline (no augmentation)} 
We refer to procedures commonly used for training FR models, which involve augmenting the training dataset solely through the application of a random horizontal flip, with a probability of $0.5$.

\noindent \paragraph{Random augmentation} \label{raug}
We evaluated the impact of enriching the FR training dataset on the FR verification accuracy by employing random augmentation techniques~\cite{randaug,unsupervisedFR}, denoted as RandAug.
The augmentation space of RandAug includes the following color and geometric transformations~\cite{unsupervisedFR}: horizontal-flipping, rotation, x-axis translation, y-axis translation, x-axis shear, y-axis shear, sharpness adjusting, auto-contrasting, adjusting contrast, solarizing, posterizing, equalizing, adjusting color, adjusting brightness, performing a resized crop, and converting to grayscale. RandAug includes two hyper-parameters, \textit{Q} and \textit{M}, to select the number of operations \textit{Q} and the magnitude \textit{M} of each transformation, respectively.
We set \textit{M} to 16 and \textit{Q} to 4, following ~\cite{unsupervisedFR}. {\color{blue}In our analysis, we first experimented with the application of a random augmentation on both subsets of the training dataset, i.e., authentic and synthetic. In a later stage, we only augmented the synthetic subset (this approach is visually depicted in Figure \ref{fig:pipeline})}.



\subsection{Model Training} \label{modeltrain}
We relied on FR models having the well-known ResNet50 architecture \cite{residualFR} as the backbone and CosFace \cite{cosface} as the loss function; the latter is defined as:
\begin{equation}
\label{eq:cosface}
\resizebox{\linewidth}{!}{$
    L_{CosFace}=\frac{1}{N}  \sum\limits_{i \in N} - log \frac{e^{s (cos(\theta_{y_i})-m)}}{ e^{s(cos(\theta_{y_i})-m)} +\sum\limits_{j=1 , j \ne y_i}^{c}  e^{s ( cos(\theta_{j}))}}
$}\end{equation}
where $c$ is the number of classes (identities), N is the batch size, $m$ is the margin penalty applied on the cosine angle $cos(\theta_{yi})$ between the feature representation $x_i$ of the sample $i$ and its class center $y_i$, $s$ is scale parameters. 
In all conducted experiments, the margin $m$ is set to 0.35 and the scale parameter $s$ to 64, following \cite{cosface}.

During the training, Stochastic Gradient Descent (SGD) is employed as an optimizer with an initial learning rate of 0.1. The learning rate is divided by 10 at 22, 30, and 40 training epochs \cite{Sface}. The models have been trained for 40 epochs with a total batch size of 512.


\subsection{Evaluation Benchmarks} \label{sec:eval}
We evaluated the trained FR models on the well-known benchmark verification pairs that accompany the following datasets: Labeled Faces in the Wild (LFW)~\cite{LFW}, AgeDB-30~\cite{agedb}, Celebrities in Frontal to Profile in the Wild (CFP-FP)~\cite{cfpfp}, Cross-Age LFW~\cite{calfw}, and Cross-Pose LFW~\cite{cplfw}. Results for all benchmarks are reported as verification accuracy in \%, thus adhering to their official evaluation protocol.



\section{Experimental Results}
\label{sec:results}
Our experiments analyzed, at first, the performance gaps between FR models trained on authentic-only and synthetic-only data created with SOTA generative methods (Section~\ref{sec:baselines}). Then, we explored {\color{blue} how combining different ratios of authentic and synthetic data can influence FR performance (Section~\ref{secRQ1a}) and} whether the combination of authentic and synthetic data can actually overcome the limitations of both data domains (Section~\ref{secRQ1}). 
Finally, we investigate the impact of augmentation on synthetic FR performance (Section~\ref{secRQ4}). 
{\color{blue} Our experiments were conducted on the GC and DC synthetic datasets. 
}

\subsection{Performance gap between FR trained on an authentic dataset and ones trained on synthetic dataset} \label{sec:baselines}
Table \ref{table:RQ1a} presents the verification accuracies on five benchmarks achieved by models trained on GC, DC and WF datasets. 
The table also presents the achieved accuracies of FR models trained on different subsets of authentic WF. 
As expected, the model trained on authentic WF (10K) outperformed the models trained on synthetic GC (10K) and DC (10K) \cite{DCFace}. The average verification accuracies achieved by the model trained on WF (10K) is 94.63\% and the achieved ones by the models trained on GC and DC are 79.38\% and 89.56\%, respectively. It can be also observed that the model trained on a subset of 3K identities of WF (average accuracies of 90.83\%) outperformed models trained on 10K identities of synthetic DC and GC datasets. This large performance gap can be reduced by combining synthetic with a small set of authentic identities as will be presented in the next section.



\subsection{Impact of combining synthetic data with a small subset of authentic data on FR performance}
\label{secRQ1a}
\import{}{RQ1a}



%
The third and fourth groups of results in Table \ref{table:RQ1a} present the achieved FR verification accuracies of FR models trained on different datasets that are built on by combining subsets of authentic and synthetic identities.
The total number of identities in each case is fixed to be equal to 10K, while the ratio of synthetic and authentic identities varies.

Combining only $1\mathrm{K}$ authentic identities with $9 \mathrm{K}$ identities from DC enhanced the FR model accuracy w.r.t. the DC synthetic baseline of $0.49\%$ on LFW, $3.90\%$ on AgeDB-30, $8.45\%$ on CFP-FP, $0.69\%$ on CA-LFW and $5.69\%$ on CP-LFW, with an overall average improvement of $3.68\%$ across datasets. Combining the same $1\mathrm{K}$ authentic identities with $9\mathrm{K}$ synthetic identities from GC resulted in even more noticeable improvements over the respective baseline: $5.31\%$ on LFW, $14.88\%$ on AgeDB-30, $20.35\%$ on CFP-FP, $7.94\%$ on CA-LFW and $17.66\%$ on CP-LFW, with an overall $12.60\%$ average improvement.



As expected, a consistent improvement was observed on all benchmarks as the ratio of involved authentic identities increased. Combining subsets of identities from GC and WF (third group of rows) steadily improved the average verification accuracy from $89.38\%$ ($9\mathrm{K}$ synthetic and $1\mathrm{K}$ authentic identities) to $94.46\%$ ($1\mathrm{K}$ synthetic and $9\mathrm{K}$ authentic identities). The same trend was also noted while combining DC subsets with the same authentic identities (fourth group of rows), with an increase in the average accuracy from $92.86\%$ to $94.68\%$.

\subsection{Impact of combining fixed numbers of synthetic identities with different subsets of authentic identities on FR performance}
\label{secRQ1}


Tables~\ref{table:RQ1casia} and~\ref{table:RQ1ms1m} present the FR verification accuracy achieved by models trained on a fixed number of synthetic identities ($10\mathrm{K}$) combined with a subset of authentic identities ($1\mathrm{K}$, $2\mathrm{K}$, $4\mathrm{K}$, $6\mathrm{K}$, $8\mathrm{K}$, or $10\mathrm{K}$) from the WF and MS-2 authentic datasets, respectively. 
Training FR models on subsets of authentic identities combined with synthetic data consistently improved the verification accuracy as the number of authentic identities in the training dataset increased. Models trained on a subset of 10K identities from WF combined with $10\mathrm{K}$ identities from GC achieved average verification accuracy improvements from $89.94\%$ ($1\mathrm{K}$ authentic identities) to $94.73\%$ ($10\mathrm{K}$ authentic identities). Similarly, training on a subset of 10K identities from WF  combined with DC improved the average verification accuracy from $93.09\%$ ($1\mathrm{K}$ authentic identities) to $95.06\%$ ($10\mathrm{K}$ authentic identities). Notably, the performance obtained by training the model on DC combined with $10\mathrm{K}$ authentic identities ($95.06\%$) was slightly worse than the one obtained by combining it with $8\mathrm{K}$ authentic identities ($95.08\%$) from WF. 

\begin{figure*}[!t]
\centering
\subfloat{
   \includegraphics[width=1.\linewidth, trim=1 1 1 1,clip]{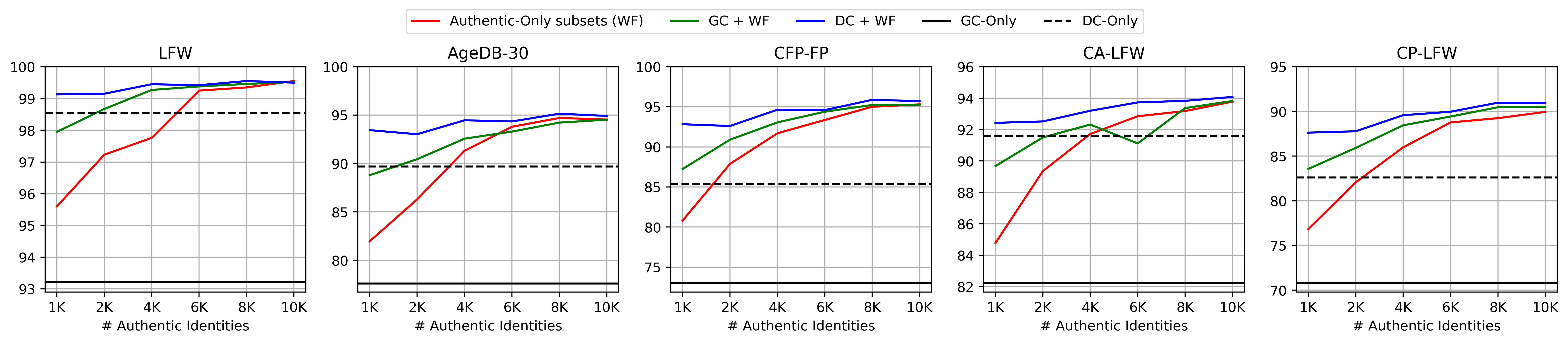}}
 \newline
\subfloat{
   \includegraphics[width=1.\linewidth, trim=1 1 1 1,clip]{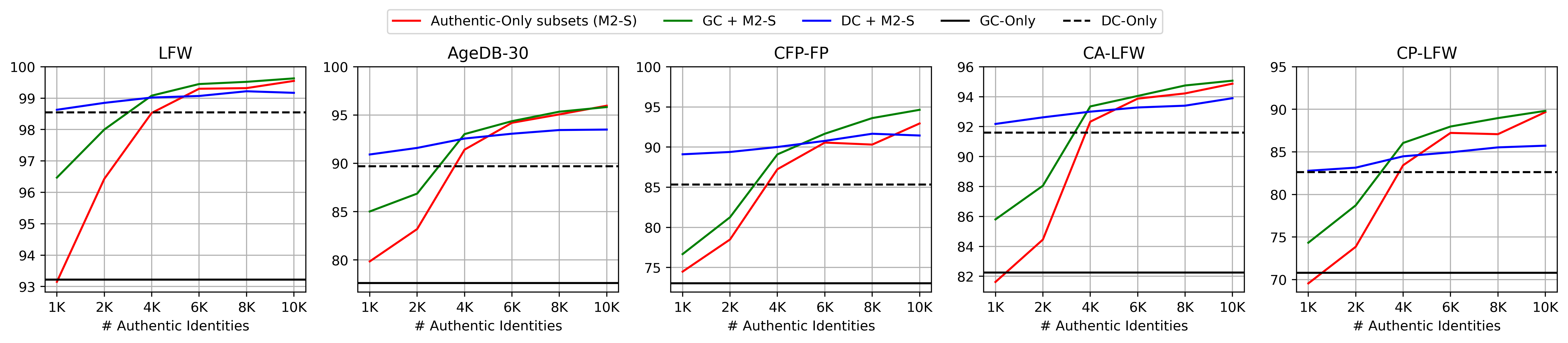}}
\caption{Average verification accuracy on testing data (y-axis) vs. number of authentic identities in training data (x-axis). The top row figures refer to authentic data sampled from WF, second row figures refer to authentic data sampled from M2-S. The verification accuracy improved by increasing the number of authentic identities in the training dataset. Also, combining synthetic with a limited subset of authentic datasets (black and green lines) improved the FR performance in comparison to the case where only a limited subset of authentic identities (red lines) is used to train FR. The number of synthetic identities in the combined dataset experiments is fixed (10K). 
The results in these plots correspond to the ones reported in Table~\ref{table:RQ1casia} Table~\ref{table:RQ1ms1m}.
}
\label{fig:comparisons}
\end{figure*}

Training FR models on M2-S combined with either GC or DC led to similar observations, reporting verification accuracy improvements from $83.66\%$ to $95.08\%$ and from $90.72\%$ to $92.75\%$, respectively. As visually emerging from Figure~\ref{fig:comparisons}, all the settings in which the models were trained on combined datasets achieved higher verification accuracy than their respective synthetic baseline, with best improvements of about $30\%$.




\import{}{RQ1_casia}

\import{}{RQ1_ms1m}

As visually illustrated in Figure~\ref{fig:comparisons}, increasing the number of authentic identities did improve, as expected, the FR performances with the best performance achieved by using 10K authentic identities, i.e. utilizing the whole authentic dataset. However, similar performances, or even higher, can be achieved by utilizing a small set of authentic identities, e.g. 4K authentic identities, combined with 10K synthetic identities from either GC or DC. 

Training the models on GC combined with authentic identity subsets from WF led to  $6.49\%$, $3.30\%$, $1.57\%$, $-0.10\%$, $0.69\%$ and $0.11\%$ improvements in terms of average verification accuracy. Similarly, training on DC combined WF subsets led to $10.83\%$, $5.02\%$, $2.81\%$, $0.85\%$, $1.26\%$  and $0.45\%$  improvements in terms of average verification accuracy. Each comparison was made by taking into consideration an authentic subset involving 1K, 2K, 4K, 6K, 8K, and 10K identities, respectively. It also emerges that, under a given amount of authentic identities, the impact of DC is more evident than that of GC.


Following the same protocol, training FR models on DC (GC) combined with authentic identity subsets from M2-S led to average verification accuracy improvements of $13.78\%$ ($4.93\%$), $9.41\%$ ($3.96\%$), $1.36\%$ ($1.69\%$), $-0.86\%$ ($0.51\%$), $-0.59\%$ ($1.33\%$), and $-1.96\%$ ($0.42\%$). These results further confirm that training on a large synthetic dataset combined with small authentic identity subsets results in a substantial increase in verification accuracy and that DC identities are more effective than those from GC. However, differently from what we observed for WF, the performance achieved by training a FR model on DC combined with 6K or more authentic identities from M2-S not only achieved lower performances compared to combining with GC identities but also adversely affected performance with respect to training only on authentic data. This observation sheds light on the importance of devising future work about methods for creating synthetic data functional to improve model accuracy, in an end-to-end manner.

\subsection{Impact of data augmentation}
\label{secRQ4}

Tables~\ref{table:RQ4gc} and~\ref{table:RQ4dc} present the achieved verification accuracies by FR models trained on different subsets of authentic and synthetic data and by augmenting the training dataset with RandAug presented in Section \ref{sec:daug}. 



The reported results demonstrated that augmenting both authentic and synthetic subsets (RAug-All settings) used to train FR models leads to overall lower average verification accuracy w.r.t. the case in which no augmentation is applied to the training data. On average, the verification accuracy of models trained on a fully augmented training dataset (with subsets from WF as authentic identities), was impacted by $-0.01\%$, $+0.23\%$, $-0.11\%$, $+0.21\%$, $-0.18\%$ and $-0.14\%$ while including synthetic identities from DC and by $+0.54\%$, $-0.43\%$, $-0.43\%$, $+0.10\%$, $-0.92\%$ and $-0.22\%$ while including synthetic identities from GC. Each comparison was made by taking into consideration combined and authentic training sets involving 1K, 2K, 4K, 6K, 8K, and 10K authentic identities, respectively.


A slight but consistent improvement, w.r.t. the baselines trained on data with no augmentation, is achieved by augmenting only the synthetic training subset (RAug-Synt settings). In this scenario, FR models trained on authentic identity subsets from WF combined with augmented synthetic samples from DC led to $-0.01\%$, $+0.49\%$, $+0.10\%$, $+0.27\%$, $-0.06\%$  and $-0.06\%$  changes in average verification accuracy. Similarly, training FR models on authentic identity subsets from WF combined with augmented synthetic samples from GC led to $+0.31\%$, $-0.30\%$, $+0.03\%$, $+0.52\%$, $0\%$, $+0.17\%$ improvements in term of average verification accuracy.

\import{}{RQ4gc}

\import{}{RQ4dc}

\import{}{MS1M_GC}

\import{}{MS1M_DC}

Given the consistent enhancements observed by augmenting only the synthetic samples in our combined training datasets, we investigated whether this behavior is confirmed across different sources of authentic identities. To address this, we extended our investigation to training FR models on synthetic data combined with authentic identities from M2-S. Results are reported in Table~\ref{table:MS1Mgc} and~\ref{table:MS1Mdc}. In these settings, it can be observed that FR models trained on augmented GC data combined with authentic identities from M2-S constantly led to slightly lower verification performances than the baselines, which were impacted by $+0.23\%$, $-2.49\%$, $-4.83\%$, $-4.09\%$, $-3.38\%$ and $-4.42\%$ while involving 1K, 2K, 4K, 6K, 8K and 10K authentic identities, respectively. 
FR models trained with augmented DC samples achieved $1.32\%$, $1.00\%$, $1.11\%$, $0.94\%$, $0.68\%$ and $0.93\%$ improvements. The consistency of our results while training FR models on augmented DC samples combined with authentic identities further reinforces our previous considerations. This points out that augmenting synthetic data is a promising direction
for improving the FR performances, in case the FR model is trained on combining subsets of synthetic and authentic data.

\section{Conclusions and Future Work}
To conclude, we connect the insights coming from individual analysis and discuss the results and future research.
Our initial analysis aimed to comprehend the performance differences between FR models exclusively trained on authentic data and those trained solely on synthetic data generated through SOTA generative methods. This investigation led to a better understanding of the impact of data source variation on model performance, under an equal number of identities across datasets. {\color{blue}Our findings align consistently with prior research~\cite{latentofromscratch,DCFace}, confirming that FR models trained on the large, full authentic dataset exhibit the highest performance, achieving an average accuracy of $94.63\%$ across the benchmarks in our experiments.


Subsequently, we investigated whether introducing a restricted number of authentic identities led to significant gains in accuracy, compared to models exclusively trained on synthetic identities. As anticipated, our study revealed a consistent improvement across all benchmarks as the number of involved authentic identities increased. FR models trained on subsets comprising both authentic and synthetic data demonstrated remarkably competitive results when compared to models exclusively trained on the entire authentic dataset. Notably, certain combinations that included DC identities even outperformed the full authentic-only baseline.


In our third analysis, we assessed whether a training dataset with a fixed number of synthetic identities and a progressively increasing number of authentic ones could attain accuracy levels, at the very least, comparable to a larger dataset comprising solely authentic identities. For both authentic datasets, all settings involving combined datasets yielded superior outcomes compared to the synthetic-only baselines, showcasing improvements of approximately $30\%$ in several cases. Moreover, training FR models on synthetic data combined with subsets of authentic identities consistently led to improvements compared to models trained exclusively on authentic subsets. As another observation, models trained on combined datasets that included synthetic data from DC consistently outperformed those involving synthetic data from GC, especially in cross-pose benchmarks.


In a last exploration, we delved into the impact of data augmentation on verification accuracy concerning the combined baselines. Throughout our experiments, we applied augmentation to two distinct sets of images: exclusively on synthetic ones and on the entire combined datasets. Notably, augmenting both synthetic and authentic samples in the combined training datasets generally resulted in inferior performance. Conversely, employing the same augmentation solely on synthetic samples led to remarkable gains in accuracy, indicating this approach as promising for enhancing accuracy when training FR models with combined datasets.}

Building upon this work, our next steps involve exploring the performance of a wider range of FR backbones and incorporating additional augmentation, sampling, and domain generalization techniques. We will also explore techniques to optimize the generation of synthetic data specifically designed for training more effective FR models when faced with (very) limited authentic data, in an end-to-end manner. Additionally, in the immediate future, we plan to extend this study with a sensitivity analysis to determine an optimal parameterization for this task, encompassing both domain generalization and data augmentation techniques. This research also suggests that augmenting limited (and likely less demographically inclusive) data with more diverse synthetic data can not only improve overall performance but also enhance fairness across demographic groups. Investigating this aspect further will be one of our next research objectives.

\balance
{\small
\bibliographystyle{ieee}
\bibliography{egbib}
}


\end{document}

%% file: RQ1a.tex
\begin{table}[t]
\begin{center}
\resizebox{\linewidth}{!}{%
\begin{tabular}{|c|c|c|c|c|c|c|c|c|}
\hline Train Data & Id/Img. & LFW $(\uparrow)$ & AgeDB-30 $(\uparrow)$ & CFP-FP $(\uparrow)$ & CA-LFW $(\uparrow)$ & CP-LFW $(\uparrow)$ & Avg $(\uparrow)$ \\
\hline
\hline GC (10K) & 10K / 50 & 93.22 & 77.60 & 73.03 & 82.25 & 70.80 & 79.38 \\
\hline DC (10K) & 10K / 50 & \textbf{98.55} & \textbf{89.70} & \textbf{85.33} & \textbf{91.60} & \textbf{82.62} & \textbf{89.56} \\
\hline
\hline WF (1K) & 1K / 45 & 95.60 & 81.97 & 80.81 & 84.78 & 76.83 &  83.99 \\
\hline WF (2K) & 2K / 45 & 97.23 & 86.30 & 87.88 & 89.36 & 82.07 & 88.57 \\
\hline WF (3K) & 3K / 45 & 98.62 & 90.32 & 90.04 & 90.83 & 84.30 & 90.82 \\
\hline WF (4K) & 4K / 45 & 97.76 & 91.32 & 91.70 & 91.72 & 85.98 & 91.69 \\
\hline WF (5K) & 5K / 45 & 99.15 & 92.75 & 93.24 & 92.40 & 87.38 & 92.98 \\
\hline WF (6K) & 6K / 46 & 99.25 & 93.80 & 93.36 & 92.85 & 88.77 & 93.61 \\
\hline WF (7K) & 7K / 46 & 99.37 & 93.58 & 94.07 & 93.13 & 89.05 & 93.84 \\
\hline WF (8K) & 8K / 46 & 99.35 & \textbf{94.71} & 95.01 & 93.18 & 89.25 & 93.90 \\
\hline WF (9K) & 9K / 46 & 99.42 & 94.32 & 95.03 & 93.73 & 89.70 & 94.44 \\
\hline WF (10K) & 10K / 46 & \textbf{99.55} & 94.55  & \textbf{95.31} & \textbf{93.78} & \textbf{89.95} & \textbf{94.63} \\
\hline 
\hline GC (1K) $\cup$ WF (9K)  & 10K / 46 & \textbf{99.43} & \textbf{94.62} & 95.10 & \textbf{93.12} & \textbf{90.02} & \textbf{94.46}  \\
\hline GC (2K) $\cup$ WF (8K) & 10K / 47 & 99.35 & 94.07 & \textbf{95.19} & 93.05 & 89.78 & 94.29  \\
\hline GC (3K) $\cup$ WF (7K) & 10K / 47 & 99.35 & 93.58 & 94.90 & 92.98 & 89.73 & 94.11  \\
\hline GC (4K) $\cup$ WF (6K) & 10K / 48 & 99.38 & 93.68 & 94.10 & 92.67 & 89.43 & 93.85  \\
\hline GC (5K) $\cup$ WF (5K) & 10K / 48 & 99.30 & 93.42 & 93.34 & 92.60 & 89.07 & 93.55  \\
\hline GC (6K) $\cup$ WF (4K) & 10K / 48 & 99.17 & 92.18 & 92.87 & 92.45 & 88.05 & 92.94  \\
\hline GC (7K) $\cup$ WF (3K) & 10K / 49 & 98.93 & 91.93 & 91.97 & 91.82 & 86.98 & 92.33  \\
\hline GC (8K) $\cup$ WF (2K) & 10K / 49 & 98.62 & 90.18 & 89.60 & 90.45 & 85.68 & 90.91  \\
\hline GC (9K) $\cup$ WF (1K) & 10K / 50 & 98.17 & 89.15 & 87.49 & 88.78 & 83.30 & 89.38  \\

\hline 
\hline DC (1K) $\cup$ WF (9K) & 10K / 46 & \textbf{99.50} & 94.38 & 95.43 & \textbf{93.63} & 90.10 & 94.61 \\
\hline DC (2K) $\cup$ WF (8K) & 10K / 47 & 99.45 & 94.65 & \textbf{95.56} & 93.55 & \textbf{90.18} & \textbf{94.68} \\
\hline DC (3K) $\cup$ WF (7K) & 10K / 47 & 99.33 & \textbf{94.75} & 95.14 & 93.52 & 90.13 & 94,57 \\
\hline DC (4K) $\cup$ WF (6K) & 10K / 48 & 99.43 & 94.28 & 95.01 & 93.42 & 90.45 & 94.52 \\
\hline DC (5K) $\cup$ WF (5K) & 10K / 48 & 99.37 & 94.25 & 94.90 & 93.33 & 89.57 & 94.28 \\
\hline DC (6K) $\cup$ WF (4K) & 10K / 48 & 99.27 & 94.10 & 94.41 & 93.33 & 89.52 & 94.13 \\
\hline DC (7K) $\cup$ WF (3K) & 10K / 49 & 99.35 & 93.55 & 94.27 & 92.87 & 89.03 & 93.81 \\
\hline DC (8K) $\cup$ WF (2K) & 10K / 49 & 99.17 & 93.27 & 93.77 & 92.68 & 88.10 & 93.40 \\
\hline DC (9K) $\cup$ WF (1K) & 10K / 50 & 99.03 & 93.20 & 92.54 & 92.23 & 87.32 & 92.86 \\
\hline
\end{tabular}
}   
\end{center}
\caption{Verification accuracy on size-fixed training sets (authentic dataset: WF). First rows group: accuracy on full synthetic datasets. Second rows group: accuracy on different subsets of authentic WF. Third and fourth rows group: accuracy on different subsets of authentic WF and synthetic GC and DC datasets, respectively. It can be noticed that the accuracy of synthetic-based models improved by adding limited subsets of authentic data. 
Highest accuracy scores for each row group in \textbf{bold}.}

\label{table:RQ1a}
\end{table}

%% file: RQ1_casia.tex
\begin{table}[!t]
\begin{center}
\resizebox{.987\linewidth}{!}{%
\begin{tabular}{|c|c|c|c|c|c|c|c|c|}
\hline Train Data & Id/Img. & LFW $(\uparrow)$ & AgeDB-30 $(\uparrow)$ & CFP-FP $(\uparrow)$ & CA-LFW $(\uparrow)$ & CP-LFW $(\uparrow)$ & Avg $(\uparrow)$ \\
\hline
\hline GC (10K) & 10K / 50 & 93.22 & 77.60 & 73.03 & 82.25 & 70.80 & 79.38 \\
\hline DC (10K) & 10K / 50 & 98.55 & 89.70 & 85.33 & 91.60 & 82.62 & 89.56 \\
\hline
\hline WF (1K) & 1K / 45 & 95.60 & 81.97 & 80.81 & 84.78 & 76.83 &  83.99 \\
\hline GC (10K) $\cup$ WF (1K) & 11K / 49 & 97.95 & 88.80 & 87.24 & 89.68 & 83.57 & 89.44 \\
\hline DC (10K) $\cup$ WF (1K) & 11K / 49 & 99.13 & 93.45 & 92.83 & 92.43 & 87.63 & \textbf{93.09} \\
\hline
\hline WF (2K) & 2K / 45 & 97.23 & 86.30 & 87.88 & 89.36 & 82.07 & 88.57 \\
\hline GC (10K) $\cup$ WF (2K) & 12K / 49 & 98.67 & 90.45 & 90.90 & 91.50 & 85.92 & 91.49 \\
\hline DC (10K) $\cup$ WF (2K) & 12K / 49 & 99.15 & 93.03 & 92.61 & 92.52 & 87.78 & \textbf{93.02} \\
\hline
\hline WF (4K) & 4K / 45 & 97.76 & 91.32 & 91.70 & 91.72 & 85.98 & 91.69 \\
\hline GC (10K) $\cup$ WF (4K) & 14K / 49 & 99.27 & 92.57 & 93.06 & 92.32 & 88.45 & 93.13 \\
\hline DC (10K) $\cup$ WF (4K) & 14K / 49 & 99.45 & 94.47 & 94.64 & 93.20 & 89.58 & \textbf{94.27} \\
\hline
\hline WF (6K) & 6K / 46 & 99.25 & 93.80 & 93.36 & 92.85 & 88.77 & 93.61 \\
\hline GC (10K) $\cup$ WF (6K) & 16K / 49 & 99.38 & 93.30 & 94.39 & 91.12 & 89.43 & 93.52 \\
\hline DC (10K) $\cup$ WF (6K) & 16K / 49 & 99.42 & 94.35 & 94.59 & 93.73 & 89.97 & \textbf{94.41} \\
\hline
\hline WF (8K) & 8K / 46 & 99.35 & 94.71 & 95.01 & 93.18 & 89.25 & 93.90 \\
\hline GC (10K) $\cup$ WF (8K) & 18K / 48 & 99.46 & 94.23 & 95.23 & 93.37 & 90.47 & 94.55 \\
\hline DC (10K) $\cup$ WF (8K) & 18K / 49 & 99.55 & 95.15 & 95.88 & 93.83 & 90.98 & \textbf{95.08} \\
\hline
\hline WF (10K) & 10K / 46 & 99.55 & 94.55 & 95.31 & 93.78 & 89.95 & 94.63 \\
\hline GC (10K) $\cup$ WF (10K) & 20K / 48 & 99.52 & 94.53 & 95.26 & 93.82 & 90.53 & 94.73 \\
\hline DC (10K) $\cup$ WF (10K) & 20K / 49 & 99.50 & 94.92 & 95.71 & 94.08 & 90.98 & \textbf{95.06} \\
\hline 
\end{tabular}
}
\end{center}
\vspace{-3mm}
\caption{Verification accuracy of models trained on synthetic datasets combined with different subsets authentic WF. First group: accuracy of models trained on synthetic GC and DC datasets, each with 10K identities. The first row in each group: accuracy of models trained by using different subsets of authentic WF only. Second and third rows in each group: accuracy of models obtained by combining different subsets of authentic WF with 10K synthetic identities from GC and DC datasets, respectively. It can be noticed that the accuracy of synthetic-based models improved by adding a limited subset of authentic data. Moreover, combining synthetic with authentic datasets (e.g., GC (10K) $\cup$ WF (10K)) led to higher accuracy, in comparison to the case where FR is only trained on authentic WF dataset/subsets.
For each row group, the highest accuracy score is highlighted in \textbf{bold}.
}
\label{table:RQ1casia}
\end{table}

%% file: RQ1_ms1m.tex
\begin{table}[!t]
\begin{center}
\resizebox{\linewidth}{!}{%
\begin{tabular}{|c|c|c|c|c|c|c|c|c|}
\hline Train Data & Id/Img. & LFW $(\uparrow)$ & AgeDB-30 $(\uparrow)$ & CFP-FP $(\uparrow)$ & CA-LFW $(\uparrow)$ & CP-LFW $(\uparrow)$ & Avg $(\uparrow)$ \\
\hline
\hline GC (10K) & 10K / 50 & 93.22 & 77.60 & 73.03 & 82.25 & 70.80 & 79.38 \\
\hline DC (10K) & 10K / 50 & 98.55 & 89.70 & 85.33 & 91.60 & 82.62 & 89.56 \\
\hline
\hline M2-S (1K) & 1K / 45 & 93.15 & 79.85 & 74.49 & 81.62 & 69.55 & 79.73 \\
\hline GC (10K) $\cup$ M2-S (1K) & 11K / 49 & 96.47 & 85.02 & 76.67 & 85.80 & 74.33 & 83.66 \\
\hline DC (10K) $\cup$ M2-S (1K) & 11K / 49 & 98.63 & 90.92 & 89.10 & 92.18 & 82.78 & \textbf{90.72} \\
\hline
\hline M2-S (2K) & 2K / 45 & 96.43 & 83.20 & 78.49 & 84.45 & 73.85 & 83.28 \\
\hline GC (10K) $\cup$ M2-S (2K) & 12K / 49 & 98.00 & 86.87 & 81.26 & 88.05 & 78.72 & 86.58 \\
\hline DC (10K) $\cup$ M2-S (2K) & 12K / 49 & 98.85 & 91.60 & 89.39 & 92.62 & 83.15 & \textbf{91.12} \\
\hline
\hline M2-S (4K) & 4K / 45 & 98.53 & 91.42 & 87.24 & 92.33 & 83.43 & 90.59 \\
\hline GC (10K) $\cup$ M2-S (4K) & 14K / 49 & 99.08 & 93.03 & 89.09 & 93.35 & 86.05 & \textbf{92.12} \\
\hline DC (10K) $\cup$ M2-S (4K) & 14K / 49 & 99.02 & 92.57 & 90.01 & 93.00 & 84.48 & 91.82 \\
\hline
\hline M2-S (6K) & 6K / 46 & 99.30 & 94.20 & 90.56 & 93.87 & 87.23 & 93.03 \\
\hline GC (10K) $\cup$ M2-S (6K) & 16K / 49 & 99.45 & 94.37 & 91.66 & 94.05 & 87.98 & \textbf{93.50} \\
\hline DC (10K) $\cup$ M2-S (6K) & 16K / 49 & 99.07 & 93.07 & 90.77 & 93.28 & 84.95 & 92.23 \\
\hline
\hline M2-S (8K) & 8K / 46 & 99.32 & 95.07 & 90.31 & 94.22 & 87.08 & 93.20 \\
\hline GC (10K) $\cup$ M2-S (8K) & 18K / 48 & 99.52 & 95.35 & 93.61 & 94.75 & 88.97 & \textbf{94.44} \\
\hline DC (10K) $\cup$ M2-S (8K) & 18K / 49 & 99.22 & 93.45 & 91.66 & 93.40 & 85.53 & 92.65 \\
\hline
\hline M2-S (10K) & 10K / 46 & 99.55 & 95.97 & 92.94 & 94.87 & 89.67 & 94.60 \\
\hline GC (10K) $\cup$ M2-S (10K) & 20K / 48 & 99.63 & 95.83 & 94.64 & 95.07 & 89.82 & \textbf{95.00} \\
\hline DC (10K) $\cup$ M2-S (10K) & 20K / 49 & 99.17 & 93.50 & 91.44 & 93.90 & 85.73 & 92.75 \\
\hline
\end{tabular}
}
\end{center}
\caption{Verification accuracy of models trained on synthetic datasets combined with different subsets authentic  M2-S. The first group: accuracy of models trained on synthetic GC and DC datasets, each with 10K identities. The first row in each group: accuracy of models trained by using different subsets of authentic M2-S only. Second and third rows in each group: accuracy of models obtained by combining different subsets of authentic M2-S with 10K synthetic identities from GC and DC datasets, respectively. It can be noticed that the accuracy of synthetic-based models improved by adding a limited subset of authentic data. Moreover, combining synthetic with authentic datasets (e.g., GC (10K) $\cup$ M2-S (10K)) led to higher accuracy, in comparison to the case where FR is only trained on authentic M2-S dataset/subsets.
For each row group, the highest accuracy score is highlighted in \textbf{bold}.}
\label{table:RQ1ms1m}
\end{table}

%% file: RQ4gc.tex
\begin{table}[!t]
\begin{center}
\resizebox{\linewidth}{!}{%
\begin{tabular}{|c|c|c|c|c|c|c|c|c|}
\hline Train Data & Method &  LFW $(\uparrow)$ & AgeDB-30 $(\uparrow)$ & CFP-FP $(\uparrow)$ & CA-LFW $(\uparrow)$ & CP-LFW $(\uparrow)$ & Avg $(\uparrow)$  \\
\hline GC (10K) $\cup$ WF (1K) & - & 97.95 &  88.80 & 87.24 & 89.68 &  83.57 & 89.44  \\ 
\hline GC (10K) $\cup$ WF (1K) & RAug-All &  98.27 &  88.87 &  89.31 & 89.35 &  83.82 &  \textbf{89.92}  \\
\hline GC (10K) $\cup$ WF (1K) & RAug-Synt &  98.25 & 88.47 &  88.76 & 89.68 & 83.47 &  \underline{89.73}  \\
\hline 
\hline GC (10K) $\cup$ WF (2K) & -     & 98.67 &  90.45 &  90.90 &  91.50 &  85.92 &  \textbf{91.49} \\ 
\hline GC (10K) $\cup$ WF (2K) & RAug-All   &  98.80 & 89.88 &  90.52 & 90.67 & 85.65 & 91.10 \\
\hline GC (10K) $\cup$ WF (2K) & RAug-Synt  &  98.83 &  90.61 & 89.60 &  91.30 &  85.75 &  \underline{91.22} \\
\hline 
\hline GC (10K) $\cup$ WF (4K) & - &  99.27 &  92.57 &  93.06 &  92.32 &  88.45 &  \underline{93.13}  \\ 
\hline GC (10K) $\cup$ WF (4K) & RAug-All & 99.02 & 92.23 & 92.69 & 92.07 & 87.67 & 92.74 \\
\hline GC (10K) $\cup$ WF (4K) & RAug-Synt &  99.10 &  92.28 &  93.03 &  92.45 &  88.92 &  \textbf{93.16} \\
\hline 
\hline GC (10K) $\cup$ WF (6K) & -     &  99.38 &  93.30 &  94.39 & 91.12 &  89.43 & 93.52 \\
\hline GC (10K) $\cup$ WF (6K) & RAug-All   &  99.37 & 92.80 & 94.33 &  92.78 & 88.81 &  \underline{93.62} \\
\hline GC (10K) $\cup$ WF (6K) & RAug-Synt  & 99.35 &  93.13 &  94.77 &  92.75 &  89.97 &  \textbf{93.79} \\
\hline 
\hline GC (10K) $\cup$ WF (8K) & - & 99.46 & 94.23 & 95.23 & 93.37 & 90.47 & \textbf{94.55}   \\ 
\hline GC (10K) $\cup$ WF (8K) & RAug-All & 99.32 & 93.25 & 94.47 & 92.35 & 89.10 & 93.70 \\
\hline GC (10K) $\cup$ WF (8K) & RAug-Synt & 99.48 & 94.25 & 95.13 & 93.48 & 90.40 & \textbf{94.55} \\
\hline 
\hline GC (10K) $\cup$ WF (10K) & -    &  99.52 &  94.53 & 95.26 &  93.82 &  90.53 &  \textbf{94.73} \\
\hline GC (10K) $\cup$ WF (10K) & RAug-All  &  99.48 & 93.97 &  95.27 & 93.53 & 90.38 & 94.53 \\
\hline GC (10K) $\cup$ WF (10K) & RAug-Synt & 99.42 &  94.37 &  95.40 &  93.57 &  90.60 &  \underline{94.67} \\
\hline
\end{tabular}}
\end{center}
\caption{{\color{blue}
Impact of augmenting the training dataset on FR verification accuracies (authentic dataset: WF; synthetic dataset: GC). For different subsets of the authentic dataset combined with the synthetic dataset, we reported the results under three settings: no augmentation, both synthetic and authentic samples are randomly augmented (RAug-All) and only synthetic samples are augmented (RAug-Synt). 
For each group of experimental settings, the best average accuracy is in \textbf{bold}.
The authentic identities are sampled from WF and the synthetic ones are from GC.}}

\label{table:RQ4gc}
\end{table}

%% file: RQ4dc.tex
\begin{table}[t]
\begin{center}
\resizebox{\linewidth}{!}{%
\begin{tabular}{|c|c|c|c|c|c|c|c|c|}
\hline Train Data & Method & LFW $(\uparrow)$ & AgeDB-30 $(\uparrow)$ & CFP-FP $(\uparrow)$ & CA-LFW $(\uparrow)$ & CP-LFW $(\uparrow)$ & Avg $(\uparrow)$ \\
\hline DC (10K) $\cup$ WF (1K) & - &  99.13 &  93.45 &  92.83 &  92.43 & 87.63 & \textbf{93.09}  \\ 
\hline DC (10K) $\cup$ WF (1K) & RAug-All &  99.15 & 93.23 &  92.81 &  92.40 &  87.82 & 93.08 \\
\hline DC (10K) $\cup$ WF (1K) & RAug-Synt &  99.13 &  93.43 & 92.80 & 92.38 &  87.70 & \textbf{93.09}  \\
\hline 
\hline DC (10K) $\cup$ WF (2K) & - & 99.15 &  93.03 & 92.61 & 92.52 & 87.78 & 93.02  \\ 
\hline DC (10K) $\cup$ WF (2K) & RAug-All &  99.18 & 93.02 &  93.57 &  92.57 &  87.83 &  \underline{93.23} \\
\hline DC (10K) $\cup$ WF (2K) & RAug-Synt &  99.25 &  93.25 &  93.48 &  92.86 &  88.46 &  \textbf{93.46} \\
\hline 
\hline DC (10K) $\cup$ WF (4K) & - &  99.45 &  94.47 &  94.64 &  93.20 &  89.58 &  \underline{94.27} \\ 
\hline DC (10K) $\cup$ WF (4K) & RAug-All & 99.33 & 94.32 &  94.86 & 92.93 & 89.37 & 94.16 \\
\hline DC (10K) $\cup$ WF (4K) & RAug-Synt &  99.37 &  94.33 & 94.60 &  93.67 &  89.83 &  \textbf{94.36} \\
\hline 
\hline DC (10K) $\cup$ WF (6K) & - &  99.42	&  94.35	& 94.59	&  93.73	& 89.97	& 94.41 \\
\hline DC (10K) $\cup$ WF (6K) & RAug-All   & 99.40	&  94.42	&  95.40	& 93.53	&  90.28 &  \underline{94.61} \\
\hline DC (10K) $\cup$ WF (6K) & RAug-Synt  &  99.50  & 94.32	&  95.06	&  93.95	&  90.48 &  \textbf{94.66} \\
\hline 
\hline DC (10K) $\cup$ WF (8K) & - & 99.55 &  95.15 &  95.88 &  93.83 & 90.98 &   \textbf{95.08}   \\ 
\hline DC (10K) $\cup$ WF (8K) & RAug-All & 99.55 &  95.02 & 95.69 & 93.78 & 90.53 & 94.91 \\
\hline DC (10K) $\cup$ WF (8K) & RAug-Synt &  99.62 & 94.77 &  95.73 &  94.02 & 90.98 &  \underline{95.02} \\
\hline 
\hline DC (10K) $\cup$ WF (10K) & -    & 99.50	&  94.92	& 95.71	&  94.08	&  90.98	&  \underline{95.04} \\
\hline DC (10K) $\cup$ WF (10K) & RAug-All  &  99.55	& 94.72	& 95.71	& 94.07	& 90.48 & 94.91 \\
\hline DC (10K) $\cup$ WF (10K) & RAug-Synt &  99.55	&  94.85	& 95.71	&  94.10	&  91.11 &  \textbf{95.06} \\
\hline
\end{tabular}}
\end{center}
\caption{{\color{blue}
Impact of augmenting the training dataset on FR verification accuracies (authentic dataset: WF; synthetic dataset: DC). For different subsets of the authentic (WF) dataset combined with the synthetic dataset (DC), we reported the results under three settings: no augmentation, both synthetic and authentic samples are randomly augmented (RAug-All) and only synthetic samples are augmented (RAug-Synt). 
For each group of experimental settings, the best average accuracy is in \textbf{bold}. Notably, augmenting only the synthetic samples led to higher verification accuracies in most of the settings, in comparison to cases where both authentic and synthetic samples are augmented.}
}
\label{table:RQ4dc}
\end{table}

%% file: MS1M_GC.tex
\begin{table}[!t]
\begin{center}
\resizebox{\linewidth}{!}{%
\begin{tabular}{|c|c|c|c|c|c|c|c|c|}
\hline Train Data & Method & LFW $(\uparrow)$ & AgeDB-30 $(\uparrow)$ & CFP-FP $(\uparrow)$ & CA-LFW $(\uparrow)$ & CP-LFW $(\uparrow)$ & Avg $(\uparrow)$ \\
\hline GC (10K) $\cup$ M2-S (1K) & - & 96.47 & 85.02 & 76.67 & 85.80 & 74.33 & 83.66  \\ 
\hline GC (10K) $\cup$ M2-S (1K) & RAug-Synt & 96.27 & 82.15 & 77.64 & 85.22 & 75.70 & 83.40  \\
\hline 
\hline GC (10K) $\cup$ M2-S (2K) & - & 98.00 & 86.87 & 81.26 & 88.05 & 78.72 & \textbf{86.58}  \\ 
\hline GC (10K) $\cup$ M2-S (2K) & RAug-Synt & 96.53 & 83.05 & 78.74 & 86.10 & 77.73 & 84.43 \\
\hline \hline GC (10K) $\cup$ M2-S (4K) & - & 99.08 & 93.03 & 89.09 & 93.35 & 86.05 & \textbf{92.12} \\ 
\hline GC (10K) $\cup$ M2-S (4K) & RAug-Synt & 97.98 & 86.60 & 83.00 & 89.35 & 81.72 & 87.73 \\
\hline 
\hline GC (10K) $\cup$ M2-S (6K) & -  & 99.45 & 94.37 & 91.66 & 94.05 & 87.98 & \textbf{93.50}  \\
\hline GC (10K) $\cup$ M2-S (6K) & RAug-Synt  & 98.63 & 88.70 & 84.80 & 91.03 & 85.40 & 89.71 \\
\hline \hline GC (10K) $\cup$ M2-S (8K) & - & 99.52 & 95.35 & 93.61 & 94.75 & 88.97 & \textbf{94.44}  \\ 
\hline GC (10K) $\cup$ M2-S (8K) & RAug-Synt & 98.95 & 91.25 & 86.46 & 92.38 & 87.30 & 91.27   \\
\hline 
\hline GC (10K) $\cup$ M2-S (10K) & - & 99.63 & 95.83 & 94.64 & 95.07 & 89.82 & \textbf{95.00}   \\
\hline GC (10K) $\cup$ M2-S (10K) & RAug-Synt & 99.33 & 87.77 & 88.00 & 91.28 & 87.70 & 90.82   \\
\hline
\end{tabular}}
\end{center}
\caption{{\color{blue} Impact of augmenting the training dataset on FR verification accuracies (authentic dataset: M2-S; synthetic dataset: GC). Verification accuracy comparison between baselines trained with combined authentic and synthetic identities (-) and the same with data augmentation applied only on synthetic samples (RAug-Synt). For each group
of results, best average accuracies are highlighted in \textbf{bold}.
The authentic identities are sampled from M2-S, the synthetic ones from GC.}}
\label{table:MS1Mgc}
\end{table}

%% file: MS1M_DC.tex
\begin{table}[!t]
\begin{center}
\resizebox{\linewidth}{!}{%
\begin{tabular}{|c|c|c|c|c|c|c|c|c|}
\hline Train Data & Method & LFW $(\uparrow)$ & AgeDB-30 $(\uparrow)$ & CFP-FP $(\uparrow)$ & CA-LFW $(\uparrow)$ & CP-LFW $(\uparrow)$ & Avg $(\uparrow)$ \\
\hline DC (10K) $\cup$ M2-S (1K) & - & 98.63 & 90.92 & 89.10 & 92.18 & 82.78 & 90.72  \\ 
\hline DC (10K) $\cup$ M2-S (1K) & RAug-Synt & 98.85 & 91.53 & 90.73 & 92.40 & 85.80 & \textbf{91.86}  \\
\hline 
\hline DC (10K) $\cup$ M2-S (2K) & - & 98.85 & 91.60 & 89.39 & 92.62 & 83.15 & 91.12  \\ 
\hline DC (10K) $\cup$ M2-S (2K) & RAug-Synt & 98.82 & 92.03 & 90.89 & 92.30 & 85.85 & \textbf{91.98} \\
\hline 
\hline DC (10K) $\cup$ M2-S (4K) & - & 99.02 & 92.57 & 90.01 & 93.00 & 84.48 & 91.82 \\ 
\hline DC (10K) $\cup$ M2-S (4K) & RAug-Synt & 99.13 & 93.05 & 92.03 & 92.90 & 86.82 & \textbf{92.79} \\
\hline 
\hline DC (10K) $\cup$ M2-S (6K) & -  & 99.07 & 93.07 & 90.77 & 93.28 & 84.95 & 92.23 \\
\hline DC (10K) $\cup$ M2-S (6K) & RAug-Synt  & 99.13 & 93.28 & 92.43 & 93.22 & 87.18 & \textbf{93.05} \\
\hline 
\hline DC (10K) $\cup$ M2-S (8K) & - & 99.22 & 93.45 & 91.66 & 93.40 & 85.53 & 92.65    \\ 
\hline DC (10K) $\cup$ M2-S (8K) & RAug-Synt & 99.27 & 93.70 & 92.74 & 93.35 & 87.20 & \textbf{93.25}   \\
\hline 
\hline DC (10K) $\cup$ M2-S (10K) & - & 99.17 & 93.50 & 91.44 & 93.90 & 85.73 & 92.75  \\
\hline DC (10K) $\cup$ M2-S (10K) & RAug-Synt & 99.38 & 93.92 & 93.17 & 93.42 & 87.95 & \textbf{93.57}   \\
\hline
\end{tabular}}
\end{center}
\vspace{-2mm}
\caption{{\color{blue} Impact of augmenting the training dataset on FR verification accuracies (authentic dataset: M2-S; synthetic dataset: DC)}. Verification accuracy comparison between baselines trained with combined authentic and synthetic identities (-) and the same with data augmentation applied only on synthetic samples (RAug-Synt). For each group
of results, best average accuracies are highlighted in \textbf{bold}.
The authentic identities are sampled from M2-S, the synthetic ones from DC.}
\vspace{1mm}
\label{table:MS1Mdc}
\end{table}